\documentclass[10pt,twocolumn,letterpaper]{article}

\usepackage{wacv}              

\usepackage{graphicx}
\usepackage{amsmath}
\usepackage{amssymb}
\usepackage{booktabs}
\usepackage{xcolor}
\usepackage{multirow}
\usepackage[pagebackref,breaklinks,colorlinks]{hyperref}

\usepackage[capitalize]{cleveref}
\crefname{section}{Sec.}{Secs.}
\Crefname{section}{Section}{Sections}
\Crefname{table}{Table}{Tables}
\crefname{table}{Tab.}{Tabs.}

\def\wacvPaperID{1307} 
\def\confName{WACV}
\def\confYear{2025}

\begin{document}

\title{Optimizing Dense Visual Predictions Through Multi-Task Coherence and Prioritization}

\author{Maxime Fontana$^1$, Michael Spratling$^2$, and Miaojing Shi$^3$\thanks{Corresponding author. }\\
$^1$Department of Informatics, King’s College London\\
$^2$Department of Behavioural and Cognitive Sciences, University of Luxembourg\\
$^3$College of Electronic and Information Engineering, Tongji University\\
{\tt\small maxime.fontana@kcl.ac.uk;~michael.spratling@uni.lu;~mshi@tongji.edu.cn}
}

\maketitle

\begin{abstract}
   Multi-Task Learning (MTL) involves the concurrent training of multiple tasks, offering notable advantages for dense prediction tasks in computer vision. MTL not only reduces training and inference time as opposed to having multiple single-task models, but also enhances task accuracy through the interaction of multiple tasks. However, existing methods face limitations. They often rely on suboptimal cross-task interactions, resulting in task-specific predictions with poor geometric and predictive coherence. In addition, many approaches use inadequate loss weighting strategies, which do not address the inherent variability in task evolution during training.
   To overcome these challenges, we propose an advanced MTL model specifically designed for dense vision tasks. Our model leverages state-of-the-art vision transformers with task-specific decoders. To enhance cross-task coherence, we introduce a trace-back method that improves both cross-task geometric and predictive features. Furthermore, we present a novel dynamic task balancing approach that projects task losses onto a common scale and prioritizes more challenging tasks during training. Extensive experiments demonstrate the superiority of our method, establishing new state-of-the-art performance across two benchmark datasets.
   The code is available at: \href{https://github.com/Klodivio355/MT-CP}{https://github.com/Klodivio355/MT-CP}
\end{abstract}

\begin{figure}[h]
  \centering
  \includegraphics[width=\linewidth]{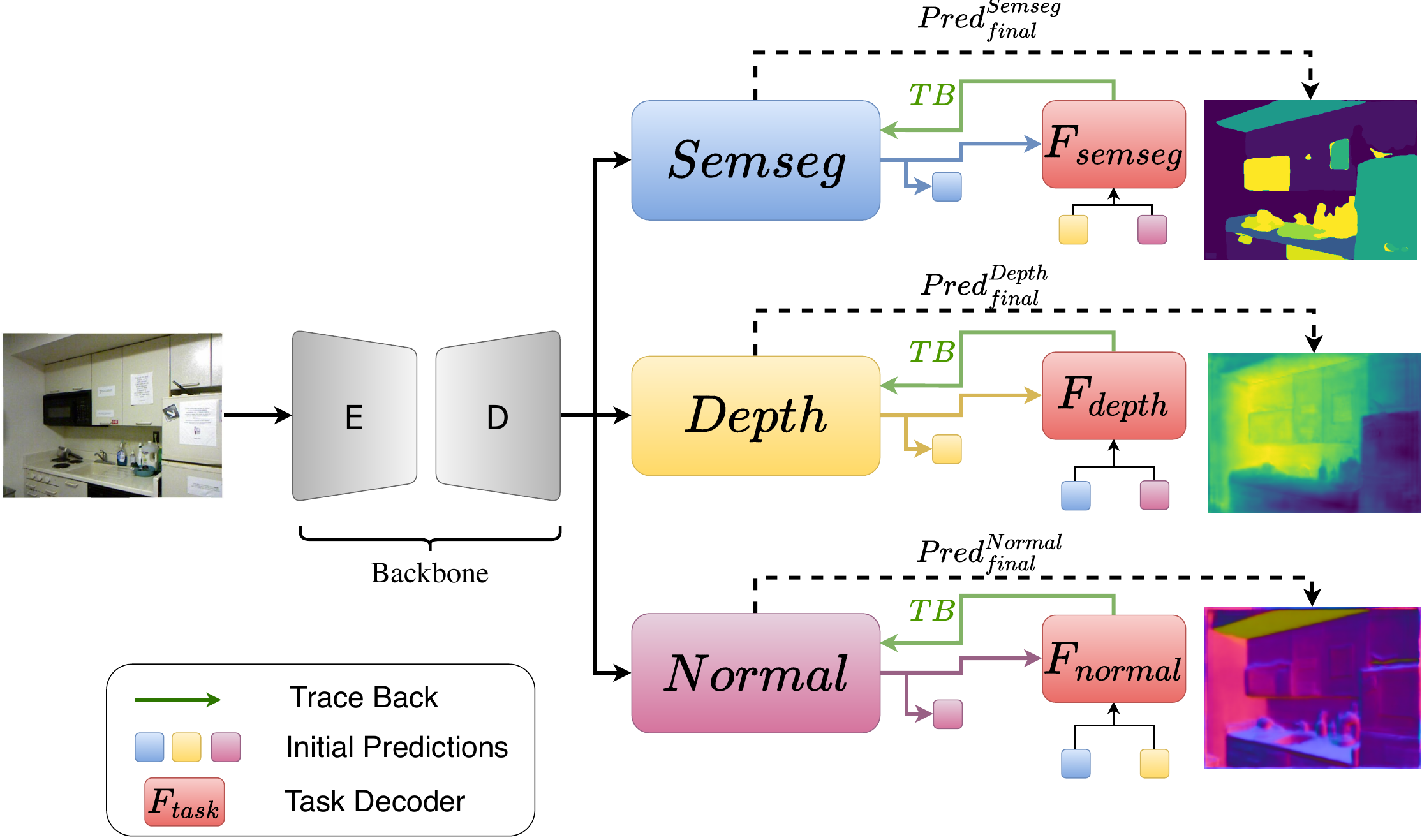}
  \caption{Our MTL framework implements cross-task coherence by tracing cross-task representations back through task-specific decoders and using them to refine the initial task predictions. The framework is optimized via a dynamic loss prioritization scheme.}
  \label{overview}
\end{figure}

\section{Introduction}
\label{sec:intro}
Dense vision tasks, which involve pixel-wise predictions, are essential to achieve a thorough understanding of scenes. These tasks encompass image segmentation \cite{semantic-decades, semantic-survey}, depth estimation \cite{depth-anything, depth-using-diff-models}, and boundary detection \cite{rethinking-boundary-detection, pushing-the-boundaries}, among others. They provide critical information that is fundamental for detailed scene analysis. Traditionally, independent models have been developed to tackle each specific task separately \cite{semantic-segmentation-sota, depth-anything, rethinking-boundary-detection}. However, there is increasing interest in developing unified models that can predict multiple tasks simultaneously. This approach, known as Multitask Learning (MTL) \cite{Ubernet, MultiMAE, MTL-dense-predic-tasks}, aims to improve the efficiency and coherence of predictions in different tasks by leveraging shared information and representations, resulting in
substantial advantages over traditional methods \cite{MTL-survey-2021, MTL-overview-2017, fontana2023}.

MTL frameworks allow interactions between tasks at various stages within the model with the aim of enhancing overall multi-task performance. On the one hand, many previous attempts consist in implementing \textit{Cross-Task Prediction Coherence}, either through distillation \cite{cross-stich, NDDR, sluice} or attention mechanisms \cite{MTFormer, crosstaskattentionmechanismdense, taskprompter}. However, these methods often result in a poor geometry consistency throughout task representations. On the other hand, we draw inspiration from \cite{Geometry-aware} to define the notion of \textit{Cross-Task Geometric Coherence}. \cite{Geometry-aware} leverages auxiliary task's geometric information to optimize the main semantic segmentation task; here, our goal is to preserve spatial relationships and geometric properties among task representations to ensure consistent geometry across all tasks. We believe that successfully solving both types of coherence as part of MTL frameworks is the key.


Another aim of MTL is for concurrent training of multiple tasks to improve parameter efficiency and create robust, transferable representations. 
However, training multiple tasks together comes with major challenges: (1) some tasks can dominate in terms of gradient magnitudes due to their task-specific loss scales, resulting in larger gradients on the shared parameters and causing hyperfocus on the larger-scaled task functions; (2) tasks do not naturally evolve at the same pace, making it crucial to control the learning pace of each task while keeping the diverse task losses on the same scale. Previous MTL approaches typically opt for one of two solutions; however, each has significant issues: (1) manually choosing weights for each task, which requires extensive trial-and-error optimization \cite{taskprompter, MTFormer, Ubernet}; (2) learning parameters, which are practically nontrivial and difficult to interpret during training \cite{MTFormer, uncertainty, auxiliary-tasks-in-MTL}. To remedy these issues, we instead propose a dynamic loss prioritization scheme which balances tasks for efficient multi-task training.


In this study, we introduce a method that explicitly addresses the aforementioned \textbf{M}ulti-\textbf{T}ask \textbf{C}oherence and \textbf{P}rioritization issues, and therefore name our method MT-CP. The MT-CP architecture distinguishes itself from existing multi-task learning (MTL) models for dense predictions in two key ways. Firstly, it ensures geometric coherence of tasks by aligning the directions of task vectors in feature spaces; then, to tackle the coherence of prediction of tasks, it propagates non-linear pixel relationships through task-specific decoders back to the shared backbone (see \cref{overview}); we name this whole procedure Trace-Back. Secondly, it employs a parameter-free loss prioritization technique that normalizes task-specific losses and dynamically emphasizes more challenging tasks throughout training.
Experiments on two benchmark datasets demonstrate that MT-CP achieves state-of-the-art performance on the NYUD-v2 \cite{NYUv2} and PASCAL-Context \cite{PASCAL} datasets.


\section{Related Work}
In this section, we review key areas relevant to our research: MTL in \cref{sec:related-mtl}, cross-task interactions for dense prediction in \cref{sec:related-interactions}  and loss weighting strategies in \cref{sec:related-interactions}. Firstly, MTL allows for simultaneous training of multiple tasks, enhancing model performance and generalization. Secondly, cross-task interactions improve the accuracy and efficiency of predictions in pixel-wise visual tasks through information sharing. Lastly, loss weighting strategies balance the contributions of different tasks, ensuring effective MTL optimization.

\subsection{Multi-Task Learning}
\label{sec:related-mtl}
Multi-Task Learning (MTL) has become increasingly popular due to its ability to leverage information across multiple tasks. MTL aims to partition features into shared and task-specific subsets. Architectures for MTL can be broadly categorized based on their approach to information sharing: (1) \emph{Soft-parameter sharing} \cite{cross-stich, progressive, NDDR, sluice} involves distinct task-specific data paths, for which each task has its own set of parameters, encouraging parameter partitioning through regularization. For example, cross-stitch networks \cite{cross-stich} originally introduce this paradigm and propose to fuse parameters by performing a linear combination of activation maps from each layer of task-specific networks. Later, MTAN \cite{MTAN} suggested the use of attention mechanisms to derive a shared set of parameters from the task-specific parameters. This framework, while computationally intensive and complex, is preferred for unrelated tasks. (2) \emph{Hard-parameter sharing} \cite{Ubernet, MTFormer, MMoE, MTAN} uses a shared backbone that is branched into lightweight task-specific decoders. This design, with its extensive feature sharing, is ideal for closely related tasks. In this work, we use a hard-parameter sharing backbone with state-of-the-art transformers, based on the idea that this simple framework is well suited for dense prediction tasks because of their related nature.

\begin{figure*}[t!]
  \centering
  \includegraphics[width=\textwidth]{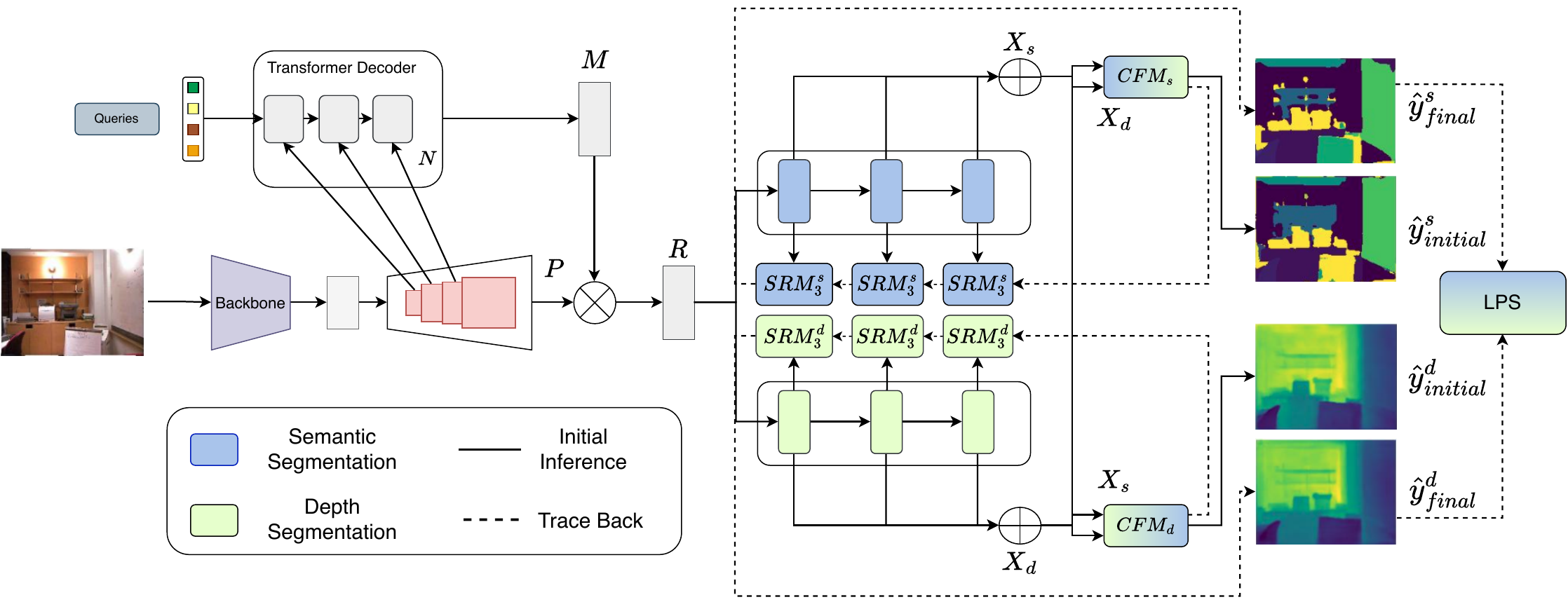}
  \caption{The proposed MT-CP model. Only two tasks are shown for clarity. The model consists of a shared set of features  extracted by a common backbone network (on the left). The model first performs a forward pass through each task-specific decoder. Next, it imposes cross-task coherence through the Coherence Fusion Module (CFM). It then traces back this cross-task representation through the Spatial Refinement Modules (SRMs) to refine an initial prediction. We optimize this model through a dynamic Loss Prioritization Scheme (LPS) which prioritizes challenging tasks throughout training.}
  \label{model}
\end{figure*}
\label{sec:model-overview}

\subsection{Cross-Task Interactions for Dense Prediction}
\label{sec:related-interactions}
Dense visual tasks in computer vision involve complex, pixel-wise, and semantically related tasks such as object detection \cite{object-detection-sota}, semantic segmentation \cite{semantic-segmentation-sota}, panoptic segmentation \cite{vlprompt}, depth estimation \cite{depth-anything}, surface normal estimation \cite{surface-normal-unsupervised} etc.. They present extremely valuable information for scene understanding. Previous MTL works have explored cross-task relationships through distillation and affinity patterns \cite{exploring-relational-context, MTI-NET, PAD-net, PAP}. Additionally, many approaches have employed visual attention mechanisms to learn non-linear relationships across tasks \cite{MTFormer, UniT,taskprompter, crosstaskattentionmechanismdense, MTAN}. However, these methods frequently fall short in explicitly identifying the high-level embeddings utilized in cross-task operations and the rationale behind their effectiveness. In contrast, we emphasize that cross-task coherence, within the context of dense visual tasks, entails maintaining both pixel-wise consistency and preserving spatial relationships across task representations. The work most closely related to ours is \cite{Geometry-aware}, which leverages geometric information from depth estimation to improve semantic segmentation. While our approach is inspired by this objective, it differs by addressing the intrinsic challenge of multi-task learning (MTL), which involves optimizing all tasks equally within a unified framework, thereby ensuring balanced performance across all tasks.

\subsection{Loss-Weighting Strategies}
\label{sec:related-loss}
In MTL training, shared parameters aggregate task-specific gradients, necessitating careful gradient design in terms of magnitudes \cite{GradNorm, gradient-vaccine} and directions \cite{gradient-surgery, gradient-vaccine}. A common strategy is to tweak task-specific loss magnitudes to indirectly manage gradient magnitudes. Many methods manually select task weights for a weighted average of gradients \cite{r-cnns, Ubernet, taskprompter}, an inefficient process requiring trial-and-error optimization. Alternatively, learning task weights during training has been explored, such as in \cite{uncertainty}, which adjusts task scalars based on uncertainty. Dynamically adjusting losses based on task difficulty is another approach, focusing on more challenging tasks during optimization \cite{dynamic-task-prio, dual-balancing, learning-to-MTL-active-sampling, lsb}. In this study, we adhere to the paradigm of dynamically adjusting the focus on challenging tasks throughout training. However, we extend this approach by also normalizing task losses to a consistent scale. Additionally, we introduce a method that enables controllable task learning paces during training. Implementing such dynamic approach enhances cross-task interactions and results in improved overall performance.

\section{Method}
In this section, we introduce the {MT-CP} Model. We present an overview of our model in \cref{sec:overview}. Next we present the technical aspects of the forward pass of our model in \cref{sec:vision-transformers}; we then illustrate how we enforce geometric coherence through the task representations in \cref{sec:coherence-fusion-module}; afterwards, we introduce in \cref{sec:traceback} how we perform the trace-back which propagates cross-task information through the task-specific decoders to help enhance predictive performance. We finally present our loss prioritization scheme in \cref{sec:lps}.

\subsection{Overview}
\label{sec:overview}

The overview method is illustrated in \cref{model}. Our MT-CP model uses a Mask2Former as a shared backbone \cite{Mask2Former} to process the RGB input. The resulting representation is then divided into task-specific heads. The representation is individually run through a pyramid transformer which provides a multi-scale representation of each task. The different scales are then concatenated by using Pyramid Feature Fusion (PFF), resulting in the task features $X_{s}$ and $X_{d}$. Subsequently, Coherence Fusion Modules (CFMs) use the aforementioned representations from both tasks to enforce pixel-wise coherence. Then, the learned embeddings are then traced back through our task decoder stages via the Spatial Refinement Modules (SRMs) attached to each stage. Throughout this prediction refinement procedure, intermediate predictions are kept and added to the MTL loss. 
Finally, predictions are obtained from the output of the final SRM module. Finally, we present a Loss Prioritization Scheme (LPS) that dynamically optimizes the learning process by prioritizing more challenging tasks. This scheme periodically updates task-specific weights based on their relative progress over a performance history. It is designed to normalize tasks on a common scale, and we further regulate task progression through the implementation of a spread parameter.

\subsection{Forward Pass}
\label{sec:vision-transformers}
\textbf{Shared Backbone.} A single input image $I \in  \mathbb{R}^{3 \times H \times W}$, is passed through a Mask2Former backbone\cite{Mask2Former}. This backbone consists of 3 elements: an encoder, a pixel decoder, and a transformer decoder. Firstly, $I$ will pass through the encoder and the pixel decoder to produce the pixel embeddings $P \in \mathbb{R}^{C \times H \times W}$. Secondly, we obtain $N$ object mask predictions from each layer in the transformer decoder, we denote those masks as $M \in \mathbb{R}^{N \times H \times W}$. We finally project the masks onto the pixel embeddings by performing matrix multiplication between the two representations: $A = P M$, then the elements in $A$ are summed over the dimension of the instance $N$, thus aggregating the contributions of each instance to produce a final representation $R \in \mathbb{R}^{N \times H \times W}$. This final representation encapsulates both the pixel-level details and the instance-level contextual information, providing a rich and informative feature map which we further utilize in the task-specific decoders. \\ 

\textbf{Task Decoders.} Given $T$ tasks, we implement task-specific decoders $F_{i=1}^{T}$. As our model is targeted towards dense prediction tasks, we choose to leverage lightweight transformer models that use Hierarchical Feature Processing (HFP) \cite{swin-transformer, pvt, PVT-v2, Deit}. As a result, we obtain the multiscale representations throughout the $K$ intermediate down-sampling stages $X_{k=1}^{K}(R_{i \in T}) \in \mathbb{R}^{(H/P) \times (W/P) \times (P^{2} \cdot C)}$, where $P$ is the hyperparameter for window size inherent to HFP transformers. Subsequently, we merge features by performing Dynamic Feature Pyramid Fusion (DFPN) \cite{pyramid-fusion}, which is a technique to integrate information across multiple scales by learning adaptive weights to selectively integrate features. The DPFN module consists of a series of Interpolation and Conv2D operations. Finally, as part of the forward pass, the coherence fusion module (CFM) uses the resulting concatenated representation to enforce geometric coherence throughout task representations. We present this method in the next section.

\subsection{Coherence Fusion Module}
\label{sec:coherence-fusion-module}
\begin{figure}[h]
  \centering
  \includegraphics[width=\linewidth]{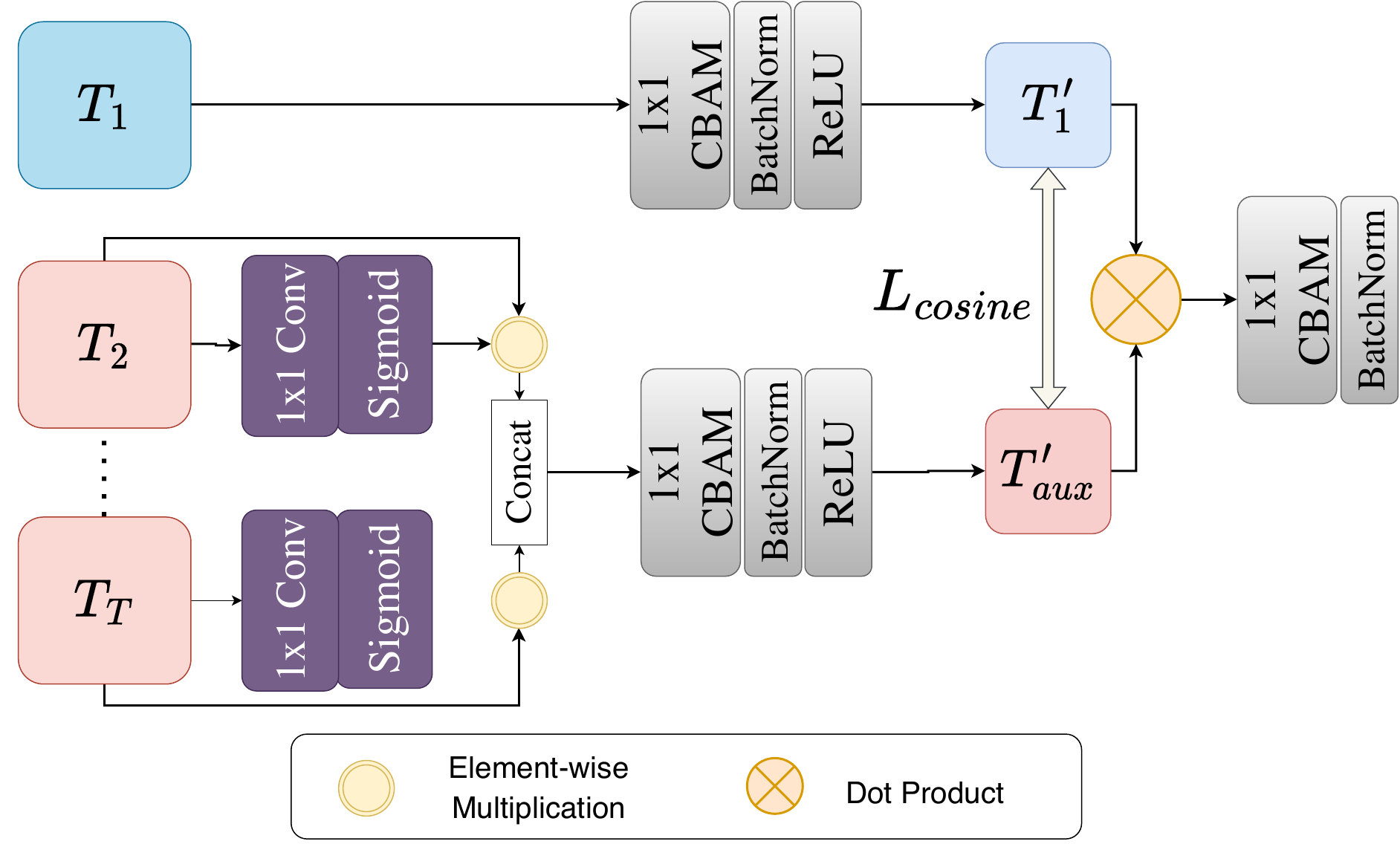}
  \caption{The coherence fusion module.}
  \label{cfm}
\end{figure}

We aim to enforce geometric coherence between tasks by using our coherence fusion module, illustrated in \cref{cfm}. CFM modules are placed at the end of each task-specific decoder and take as input (1) a main task representation $X_{T_{1}}$ and (2) a gated concatenation of all other (auxiliary) task representations $X_{T_{2...T}}$. Specifically, we design the gates as sigmoid-activated pixel-wise convolutions, which we later multiply element-wise with the original representations. We then concatenate these representations and denote the resulting representation as $X_{T_{aux}}$. Subsequently, $X_{T_{1}}$ and $X_{T_{aux}}$ are individually processed by lightweight learnable convolution-based modules that consist of a $1 \times 1$ Convolutional Block Attention Module (CBAM) \cite{cbam}, followed by a batch normalization and a ReLU activation function. We use the notation $X_{T'_{1}}$ and $X_{T'_{aux}}$ to describe the resulting representations. Then, we design two strategies to enforce geometric coherence to help enhance the main task. Firstly, we minimize the cosine distance between $X_{T'_{1}}$ and $X_{T'_{aux}}$, the cosine distance ensures that the vectors in each representation are attracted together towards the same direction. This conceptually helps ensure that the geometric structure (e.g., edges, boundaries) of the scenes is similarly captured in both representations. Secondly, the features are merged via matrix multiplication. This conceptually ensures that not only are the structural features aligned but also vector magnitudes help maintain consistency as using matrix multiplication to project onto a common space serves this purpose. Finally, the resulting representation is passed through a 1x1 CBAM \cite{cbam} and batch normalization. We note the output of the CFM : $H_{i \in T}$, $T$ being the set of tasks.

\subsection{Prediction Refinement via Trace-Back}
\label{sec:traceback}

\begin{figure}[h]
  \centering
  \includegraphics[width=\linewidth]{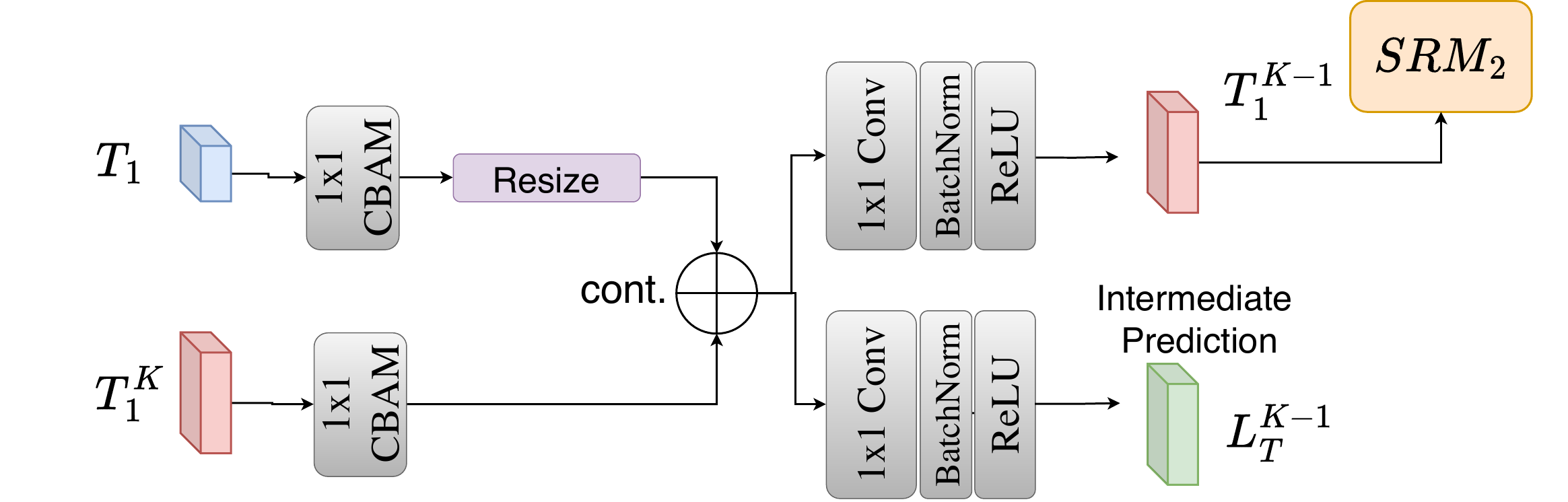}
  \caption{The spatial refinement module used to trace back cross-task embeddings.}
  \label{srm}
\end{figure}

We further leverage pixel-wise cross-task relationships for better cross-task prediction coherence. Specifically, we choose to trace back our cross-task representation from our initial representation $H_{i \in T}$ through the associated task-specific decoder blocks. This trace-back is performed through the use of the spatial refinement module, illustrated in \cref{srm}. 
Specifically, to give an example, we design our SRM to recursively propagate the cross-task representation $T_{1}$ back through Task 1 and the block scales $K$ in a bottom-up manner. Therefore, our first SRM takes as input $T_{1}$ and $T^{K}_{1}$. Subsequently, the CBAM \cite{cbam} convolutions are run to learn discriminative characteristics to better suit task 1. $T_{1}$ is resized to match the size of $T^{K}_{1}$. The learned features are then concatenated along the channel dimension before parallel and independent lightweight learnable modules consisting of pixelwise convolution, batch normalization, and the ReLU activation function are applied to produce the input $T_{1}^{K-1}$ to the next SRM module, which will also take as input $T^{K}_{2}$ and so on... In addition, as proposed by \cite{Geometry-aware}, we retain intermediate task-specific predictions to contribute to the MTL loss that aims to further improve discriminative power. 

\subsection{Loss Prioritization Scheme}
\label{sec:lps}
This section describes the design of our Loss Prioritization Scheme (LPS) to tackle the loss imbalance problem. To further improve performance by enhancing cross-task interactions throughout training, we believe that difficult tasks should not only be prioritized but also projected onto a similar scale. To this end, we first introduce the minimization objective inherent to MTL training and explain why designing an LPS is central to our challenge. Then, we introduce how we project losses onto a similar scale. Finally, we present our LPS algorithm and present our MTL loss.
\\
\indent \textbf{Objective and Problem.}
We describe a MTL objective, as finding a set of parameters $\theta^{*}$ such as :
\begin{equation} \label{eq:mtl-objective}
\theta^{*} = \arg \min_{\theta^{1}, ... ,\theta^{T}} (L^{1}(\theta^{sh},\theta^{1}), ... , L^{T}(\theta^{sh},\theta^{T})),
\end{equation}
where task specific losses $L_{i=1}^{T}$ take as parameters both the shared parameters $\theta^{sh}$ and task-specific parameters $\theta^{i \in T}$, where $T$ is the set of tasks. To achieve this objective, existing MTL methods weigh the tasks according to pre-defined weights $w_{i}$ as follows:
\begin{equation} \label{eq:basic-mtl}
L_{MTL} = \sum_{i=1}^{T} w_{i} L_{i}, 
\end{equation}
when $w_{i} = \frac{1}{T} \;\forall i$,  this is an Equal Weighting (EW) loss scheme. Otherwise, if the weights have different values, we consider this to be the Manual Annotation (MA) loss scheme. However, both loss schemes have drawbacks, EW completely overlooks the different scales, leading to a domination of the semantic segmentation task on NYUD-v2 \cite{NYUv2} for instance. This leads to undesirable overall performance caused by the faster convergence of the segmentation task. One may be interested in having tasks trained at a similar pace. Therefore, some works have chosen to perform MA to compensate for that scale difference \cite{Ubernet, taskprompter}, however, this requires a lot of trial-and-error tuning and it is also heavily dependent on the model complexity. 
We stress therefore the importance of both (1) projecting tasks onto a similar scale, (2) dynamically prioritising the more challenging tasks. 
\\
\indent \textbf{Loss Scale Projection.}
Similar to previous work \cite{uncertainty, auxiliary-tasks-in-MTL, dual-balancing}, we choose to project tasks onto a similar scale by using the $\log$ transformation. Precisely, we choose to formulate our overall objective as follows:
\begin{equation} \label{eq:log-mtl}
L_{Log-MTL} = \sum_{i=1}^{T} \log(1+w_{i}) L_{i},
\end{equation}
where the $\log(1+w_{i})$ is necessary to avoid values for $w_{i} \in [0,1]$ leading negative weights, therefore leading to negative loss values. This scaling method has the effect to remove the scale imbalance problem.
\\
\indent \textbf{Task Prioritization.}
In addition to projecting tasks onto a similar scale through the log transformation, dynamically adjusting the learning of some tasks over others might improve the learned cross-task relationships in our CFM module. We choose to prioritise challenging tasks, which might change over training to further smooth out the training of tasks and increase overall performance. We periodically adjust the rate of tasks, at each epoch $n$. For the sake of simplicity, we denote $L_{i}$ to be the loss for a task $i \in T$ according to \cref{eq:log-mtl}, where $T$ is the set of tasks. Moreover, we define the ratio to which a task $i$ contributes to the overall loss as $\frac{L_{i}^{n}}{L^{n}}$. 
We then define an arbitrary task history length $H$. Then, we dynamically adjust our task-specific weights $\Tilde{w}^{n}_{i}$ over our history size $H$ such that:
\begin{equation} \label{eq:update}
\Tilde{w}^{n}_{i} = \frac{\prod_{k=1}^{H} \frac{L_{i}^{n-k+1}}{L_{i}^{n-k}}}{\prod_{k=1}^{H} \frac{L^{n-k+1}}{L^{n-k}}}.
\end{equation}
As a result, the weights $\Tilde{w}_{i}^{n}$ indicate whether the task-specific loss decreases quickly $(\Tilde{w}_{i}^{n}<1)$ or slowly  $(\Tilde{w}_{i}^{n}>1)$). This indicates whether a task is easy or difficult, therefore assigning more weight to the slower or difficult task, respectively.

\textbf{Controlling Spread.} As our experiments show that weights tend to be different at start and then close together as training continues. We implement a penalty term that encourages the spread of the weights around their mean. 
Firstly, let us consider the mean of the weights $\mu_{i}^{n}$ for a given epoch $n$ and task $i$.
Secondly, we calculate the deviations from the mean as follows : 
\begin{equation} \label{eq:deviations}
\sigma^{n}_{i} = w^{n}_{i} - \mu_{i}^{n} 
\end{equation}
Finally, we design a hyper-parameter $\kappa$ to scale the deviations $\sigma^{n}_{i}$ to update our weights such as : 
\begin{equation} \label{eq:weights}
w'^{n}_{i} = \mu_{i}^{n} + \kappa \sigma^{n}_{i} 
\end{equation}
As a result, $\kappa$ is a hyper parameter which controls the convergence of task losses by controlling the spread of our task-specific weights. Increasing $\kappa$ will lead to a higher penalty in the weights normalization.

\textbf{MTL Loss.} We summarise our overall MTL loss used for training. In addition to $L_{Log-MTL}$ defined in \cref{eq:log-mtl}, we keep track of intermediate task-specific predictions to further improve the performance. Formally, our MTL loss, for a given epoch $n$ can be formulated as below:
\begin{equation} \label{eq:lps-eq}
\begin{aligned}
L^{n}_{LPS} = L_{Log-MTL}(w_{n},L_{n}) + \sum_{i=1}^{T} \sum_{j=1}^{K} L_{i}^{j} \\
\textrm{s.t. } w^{*} = LPS(w,\kappa)
\end{aligned}
\end{equation}
where $K$ is the number of down-sampling stages in our task-specific decoder, and $w_{n}$ and $L_{n}$ represent the list of weights and losses for all tasks, for a given epoch $n$, respectively.

\section{Experiments}
\subsection{Datasets}
We apply our model on two widely used MTL datasets. \\ 
\textbf{NYUD-v2.}\cite{NYUv2} This dataset comprises 1449 labeled images drawn from indoor scene videos for which each pixel is annotated with a depth value and an object class. Additionally, there are 407,024 unlabeled images which contain RGB, depth and accelerometer data, rendering this dataset useful for real-time applications as well. This dataset comprises 3 different tasks: Semantic Segmentation, Monocular Depth Estimation and Surface Normal Estimation. \\
\textbf{Pascal-Context.} \cite{PASCAL} A dataset of 1464 of regular object-centered scenes. 
This dataset comprises 3 different tasks: Semantic Segmentation, Human Part Parsing which is a type of semantic segmentation task where objects are defined as body parts, and Saliency Detection.

\subsection{Implementation}
\begin{itemize}
  \item \textit{Semantic Segmentation / Human Parsing:} To train this task, we choose to employ the Cross Entropy loss. 
  To evaluate this task, we choose to leverage the mean Intersection over Union (mIoU).
  \item \textit{Monocular Depth Estimation:} We  leverage the L1 loss for training. We report the results of depth estimation using the Root Mean Squared Error (RMSE) value.
  \item \textit{Surface Normal Estimation:} Similarly, we choose to use the L1 loss with normalisation during training. We evaluate this task by using the mean Error (mErr).
  \item \textit{Saliency Detection:} We leverage the Balanced Cross Entropy loss function. We also adopt the maximum F-measure (maxF) to evaluate saliency detection results.
\end{itemize}

\textbf{Backbone.} We fine-tune our backbone which is a Mask2Former \cite{Mask2Former} pre-trained on the ADE20K dataset \cite{ade20k} on the semantic segmentation task. This backbone uses a small Swin transformer encoder \cite{swin-transformer}. This backbone network channel size is 256 which operates on image sizes of $(480, 640)$ for NYUD-v2 \cite{NYUv2} and $(512, 512)$ for Pascal-Context \cite{PASCAL}. \\

\textbf{Task Decoders.} Furthermore, we design lightweight task-specific decoders consisting of 3 down-sampling stages with a lightweight configuration of $(2,2,2)$ blocks per head with depth $(1,2,1)$. \\

\textbf{Network Parameters.} We validate and train our model on a NVIDIA A100 GPU. We choose to use a learning rate of $5 \times 10^{-5}$ on a batch size of $2$. We also choose an Adam optimizer with weight decay \cite{weight_decay} with a weight decay value of $1 \times 10^{-4}$. We empirically choose the value of $\kappa$ to be 2.5. Similarly, we choose the history length to be $H=3$.
\\

\subsection{Comparison with State-of-the-art}
\begin{table}[t!]
\centering
\caption{Comparison to SOTA methods on NYUD-v2 \cite{NYUv2}.}
\resizebox{\columnwidth}{!}{
    \huge
    \renewcommand{\arraystretch}{1.2} 
    \begin{tabular}{lccc}
    \toprule
    \textbf{Model} & \textbf{Semseg (mIoU) $\uparrow$} & \textbf{Depth (RMSE) $\downarrow$} & \textbf{Normal (mErr) $\downarrow$} \\
    \midrule
    Cross-Stitch \cite{cross-stich} & 36.34 & 0.6290 & 20.88 \\
    PAP \cite{PAP} & 36.72 & 0.6178 & 20.82 \\
    PSD \cite{PSD} & 36.69 & 0.6246 & 20.87 \\
    PAD-Net \cite{PAD-net} & 36.61 & 0.6270 & 20.85 \\
    MTI-Net \cite{MTI-NET} & 45.97 & 0.5365 & 20.27 \\
    InvPT \cite{invPT} & 53.56 & 0.5183 & 19.04 \\
    DeMT \cite{DeMT} & 51.50 & 0.5474 & 20.02 \\
    MLoRE \cite{2024multitaskdensepredictionmixture} & 55.96 & 0.5076 & \textbf{18.33} \\
    Bi-MTDP \cite{bi-mtdp} & 54.86 & 0.5150 & 19.50 \\
    \midrule
    STL$_{\text{Semseg}}$ & 53.20 & - & - \\
    STL$_{\text{Depth}}$ & - & 0.4923 & - \\
    STL$_{\text{Normal}}$ & - & - & 19.22 \\
    MT-CP & \textbf{56.25} & \textbf{0.4316} & 18.60 \\
    \bottomrule
    \end{tabular}
}
\label{tab:nyud}
\end{table}

\begin{table}[t!]
\centering
\caption{Comparison to SOTA methods on Pascal-Context \cite{PASCAL}.}
\resizebox{\linewidth}{!}{
    \huge
    \renewcommand{\arraystretch}{1.2} 
    \begin{tabular}{lccc}
    \toprule
    \textbf{Model} & \textbf{Semseg (mIoU) $\uparrow$} & \textbf{Parsing (mIoU) $\uparrow$} & \textbf{Saliency (maxF) $\uparrow$} \\
    \midrule
    Cross-Stitch \cite{cross-stich} & 63.28 & 60.21 & 65.13 \\
    PAD-Net \cite{PAD-net} & 60.12 & 60.70 & 67.20 \\
    MTI-Net \cite{MTI-NET} & 61.70 & 60.18 & 84.78 \\
    InvPT \cite{invPT} & 79.03 & 67.71 & 84.81 \\
    MTFormer \cite{MTFormer} & 74.15 & 64.89 & 67.71 \\
    DeMT \cite{DeMT} & 75.33 & 63.11 & 83.42 \\
    Bi-MTDP \cite{bi-mtdp} & 79.83 & 68.17 & \textbf{84.92} \\
    \midrule
    STL$_{\text{Semseg}}$ & 75.10 & - & - \\
    STL$_{\text{Parsing}}$ & - & 68.29 & - \\
    STL$_{\text{Saliency}}$ & - & - & 82.22 \\
    MT-CP & \textbf{79.96} & \textbf{69.13} & 84.20 \\
    \bottomrule
    \end{tabular}
}
\label{tab:pascal}
\end{table}

In this section, we compare our method with several state-of-the-art (SOTA) models on two benchmark datasets: NYUD-v2 \cite{NYUv2} and Pascal-Context \cite{PASCAL}. Our comparison focuses on multi-task learning performance, using only RGB input, across different tasks within these datasets.

\textbf{NYUD-v2. \cite{NYUv2}} \cref{tab:nyud} presents the performance comparison of various SOTA methods on the NYUD-v2 dataset for three tasks: semantic segmentation (Semseg), depth estimation (Depth), and surface normal estimation (Normal). Our method achieves the best performance in semantic segmentation and depth estimation, with mIoU of 56.25 and RMSE of 0.4316, respectively. Furthermore, our method shows competitive performance in normal estimation with an mErr of 18.60. Compared to the previous method with the best performance, MLoRE \cite{2024multitaskdensepredictionmixture}, our model exceeds it in both Semseg and Depth tasks. Specifically, our model improves the mIoU from 55.96 to 56.25 and reduces the RMSE from 0.5076 to 0.4316, demonstrating significant advancements. Although MLoRE \cite{2024multitaskdensepredictionmixture} achieves the best mErr of 18.33 in Normal estimation, the performance of our method is close to an mErr of 18.60.

\textbf{Pascal-Context. \cite{PASCAL}}  \cref{tab:pascal} showcases the comparison on the Pascal-Context dataset, focusing on semantic segmentation (Semseg), human part parsing (Parsing), and saliency detection (Saliency). Our approach yields top-tier results in parsing and semseg, achieving the highest mIoU of 69.13 and 79.96 respectively. In saliency detection, our method scores a maxF of 84.20, closely trailing the leading score of 84.94 by Bi-MTDP \cite{bi-mtdp}.

\begin{figure*}[t!]
  \centering
  \includegraphics[width=0.85\textwidth]{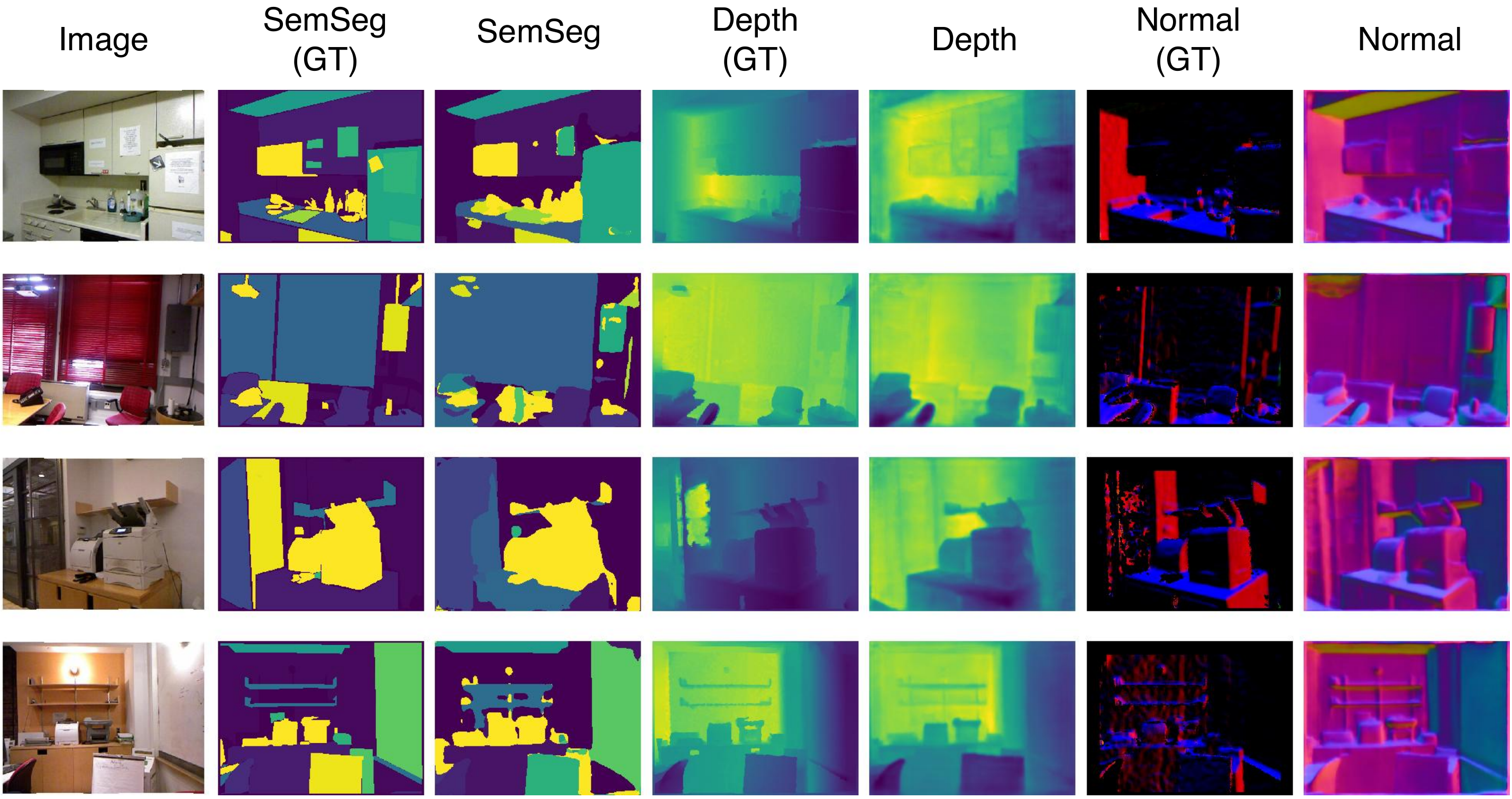}
  \caption{Visualisations of predictions on NYUD-v2 \cite{NYUv2}.}
  \label{nyud_vis}
\end{figure*}

Overall, our approach demonstrates substantial improvements and competitive results across both datasets, establishing it as a strong contender in the multi-task learning domain. These results highlight the effectiveness of both our model architecture and our loss-balancing strategy in enhancing performance across diverse tasks.
Some visualizations of our model predictions on this dataset are shown in \cref{nyud_vis}. \\

\begin{figure*}[t!]
  \centering
\includegraphics[width=\textwidth]{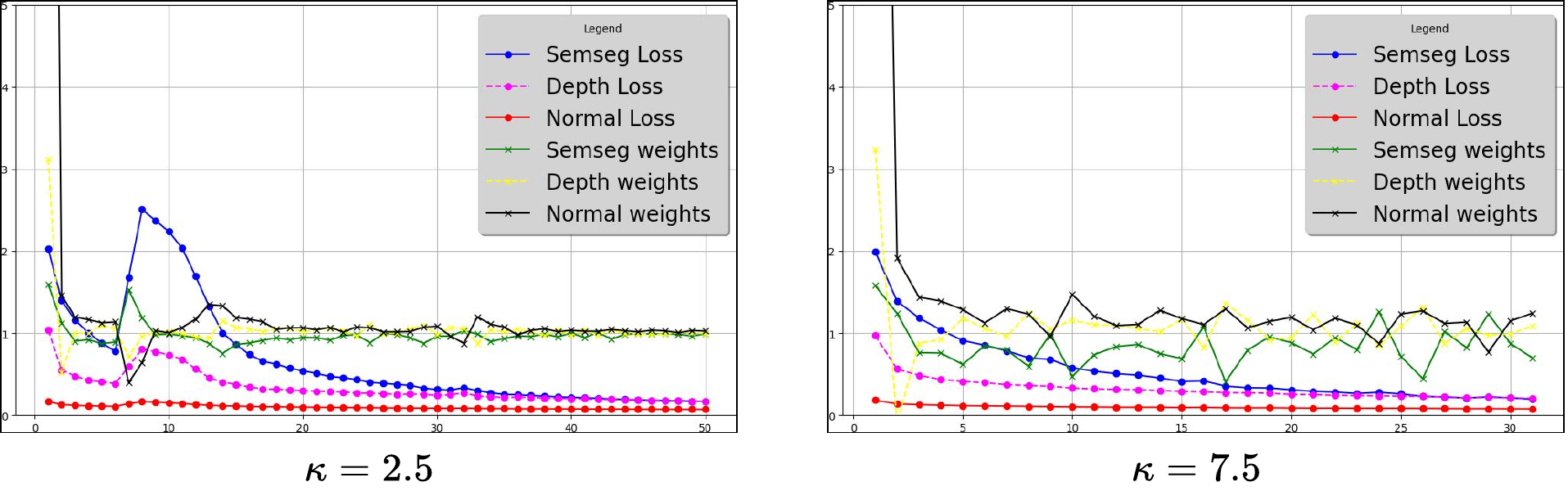}
  \caption{Variation of the spread value $\kappa$ on our Loss Prioritization Scheme (LPS).}
  \label{loss-kappa}
\end{figure*}

\subsection{Ablation Analysis}

\begin{table}[t!]
\centering
\caption{Hierarchical Ablation on NYUD-v2 \cite{NYUv2}}
\resizebox{\columnwidth}{!}{
    \Huge
    \renewcommand{\arraystretch}{1.4} 
    \begin{tabular}{lccc}
    \toprule
    \textbf{Model} & \textbf{Semseg (mIoU) $\uparrow$} & \textbf{Depth (RMSE) $\downarrow$} & \textbf{Normal (mErr) $\downarrow$} \\
    \midrule
    MT-CP & \textbf{56.25} & \textbf{0.4316} & \textbf{18.60} \\
    \midrule
    MT-CP w/o CFM & 52.78 & 0.4803 & 19.20 \\
    MT-CP w/o SRM & 55.12 & 0.4561 & 18.95 \\
    MT-CP w/o CFM \& SRM & 54.02 & 0.5025 & 20.50 \\
    \midrule
    STL$_{\text{Semseg}}$ & 53.20 & - & - \\
    STL$_{\text{Depth}}$ & - & 0.4923 & - \\
    STL$_{\text{Normal}}$ & - & - & 19.22 \\
    \bottomrule
    \end{tabular}
}
\label{tab:ablation_nyud}
\end{table}

\begin{table}[t!]
\centering
\caption{Loss Scheme Study on NYUD-v2 \cite{NYUv2}}
\resizebox{\columnwidth}{!}{%
    \Huge
    \renewcommand{\arraystretch}{1.4} 
    \begin{tabular}{lccc}
    \toprule
    \textbf{Model} & \textbf{Semseg (mIoU) $\uparrow$} & \textbf{Depth (RMSE) $\downarrow$} & \textbf{Normal (mErr) $\downarrow$} \\
    \midrule
    MT-CP (LPS) & \textbf{56.25} & \textbf{0.4316} & \textbf{18.60} \\
    \midrule
    MT-CP (w/ EW) & 49.23 & 0.5519 & 23.80 \\
    MT-CP (w/ Log Smoothing) & 55.25 & 0.4516 & 20.60 \\
    MT-CP (w/ Loss Prioritization) & 54.50 & 0.4823 & 20.32 \\
    \midrule
    STL$_{\text{Semseg}}$ & 53.20 & - & - \\
    STL$_{\text{Depth}}$ & - & 0.4923 & - \\
    STL$_{\text{Normal}}$ & - & - & 19.22 \\
    \bottomrule
    \end{tabular}%
}
\label{tab:loss_nyud}
\end{table}

\textbf{MT-CP Architecture.} \cref{tab:ablation_nyud} illustrates the impact of key architectural components, CFM (Coherence Fusion Module) and SRM (Spatial Refinement Module), on the performance of our MT-CP model on the NYUD-v2 dataset. The complete MT-CP model, with both CFM and SRM, delivers the best results across all metrics, indicating their crucial role in the architecture. Removing CFM results in a noticeable decline in performance, particularly in semantic segmentation (mIoU drops to 52.78) and depth estimation (RMSE increases to 0.4803), highlighting its importance in feature integration to enhance geometric coherence between tasks. The absence of SRM also degrades performance, though less severely.
suggesting its role in refining spatial features for better cross-task predictive coherence. The combined removal of both CFM and SRM leads to the most significant performance drop, 
demonstrating the synergistic effect of these components in the MT-CP architecture. This ablation study confirms the critical contributions of CFM and SRM to the overall performance and robustness of the model.

\textbf{LPS.} \cref{tab:loss_nyud} presents a comparative study of various loss schemes on the NYUD-v2 dataset \cite{NYUv2}. 
MT-CP, using the Loss Prioritization Scheme (LPS), achieves superior results on all tasks.
In contrast, the Equal Weights (EW) scheme significantly underperforms, demonstrating the necessity of a balanced loss approach.
The log smoothing scheme, which consists of a simple log transform as presented in \cref{sec:lps}, offers notable improvements, yet falls short of LPS, while the Loss Prioritization (without log smoothing) configuration, although effective, does not match the consistency between tasks achieved by LPS. This analysis underscores the effectiveness of LPS in enhancing multi-task learning performance by appropriately balancing task contributions, hence resulting in a better optimization and learning of cross-task information.

\textbf{Varying $\kappa$.} We illustrate the effect of varying the hyper-parameter $\kappa$ in \cref{loss-kappa}. We show the effect of the heuristic values of $\kappa = 2.5$ and $\kappa = 7.5$ on our MTL optimization. For each given epoch, we notice that if a task-specific loss decreases slowly, the respective weights go up. We also show how a higher value of $\kappa = 7.5$ acts a stronger penalty, as opposed to $\kappa = 2.5$ to the convergence of the weights.

\section{Conclusion}
This paper introduces MT-CP, a multi-task learning model designed for dense prediction tasks. MT-CP effectively leverages pixel-wise cross-task information through each task-specific decoder, ensuring coherent predictions in both semantic and geometric contexts. Furthermore, we propose a loss prioritization scheme that dynamically focuses on more challenging tasks during training. Experimental results on two benchmark datasets demonstrate the superior performance of MT-CP, surpassing current state-of-the-art methods in certain tasks and maintaining competitive results in others. 

{\small
\bibliographystyle{ieee_fullname}
\bibliography{Bibliography}
}

\end{document}